\documentclass{article}

\usepackage{microtype}
\usepackage{graphicx}
\usepackage{subcaption}
\usepackage{booktabs} 

\usepackage{hyperref}

\usepackage[preprint]{icml2026}

\usepackage{amsmath}
\usepackage{amssymb}
\usepackage{mathtools}
\usepackage{amsthm}

\usepackage[capitalize,noabbrev]{cleveref}

\theoremstyle{plain}

\theoremstyle{definition}

\theoremstyle{remark}

\usepackage[textsize=tiny]{todonotes}

\usepackage{hyperref}
\usepackage{url}
\usepackage{algorithm}
\usepackage{algorithmic}

\usepackage{caption} 
\usepackage{booktabs}
\usepackage{tikz}
\usepackage{threeparttable}  
\usepackage{makecell}
\usepackage{multirow}
\usepackage{multicol}
\usetikzlibrary{arrows.meta, positioning, matrix, decorations.pathreplacing}
\tikzset{
  layer/.style = {
    draw,
    thick,
    minimum width=2cm,
    minimum height=0.5cm,
    align=center,
    rounded corners=2pt,
    anchor=center
  },
  arrow/.style = {
    -{Latex[length=3mm,width=2mm]},
    thick
      },
  brace label/.style = {
    font=\footnotesize,
    midway,
    align=center
  }
}

\icmltitlerunning{Post-Training Second-order Compression for Spiking Neural Networks}

\begin{document}

\twocolumn[
  \icmltitle{Post-Training Second-order Compression for Spiking Neural Networks}



  \icmlsetsymbol{equal}{*}

  \begin{icmlauthorlist}
    \icmlauthor{Lianfeng Shi}{bristol}
    \icmlauthor{Ao Li}{bristol}
    \icmlauthor{Benjamin Ward-Cherrier}{bristol}
  \end{icmlauthorlist}

  \icmlaffiliation{bristol}{School of Engineering Mathematics and Technology, University of Bristol, England}

  \icmlcorrespondingauthor{Benjamin Ward-Cherrier}{b.ward-cherrier@bristol.ac.uk}

  \icmlkeywords{Machine Learning, ICML}

  \vskip 0.3in
]



\printAffiliationsAndNotice{}  

\begin{abstract}
Spiking Neural Networks (SNNs) have emerged as a new generation of energy-efficient neural networks suitable for implementation on neuromorphic hardware. As neuromorphic hardware has limited memory and computational resources, parameter pruning and quantization have recently been explored to improve the efficiency of SNNs. State-of-the-art SNN pruning/quantization methods employ multiple compression and training iterations, increasing the cost for pre-trained or very large SNNs. In this paper, we propose a novel one-shot post-training compression framework, Spiking Brain Compression (SBC), that extends the classical Optimal Brain Surgeon method to SNNs. SBC replaces the current-based objective found in the common layer-wise compression method with a spike-train-based objective whose Hessian is cheaply computable, allowing a single backward pass to compress parameters and analytically rescale the rest. Applying SBC to SNN pruning and quantization across event-based and static datasets (up to ImageNet), including SEW-ResNet152 and spike-driven Transformers, we achieve state-of-the-art one-shot post-training compression for SNNs, with single- to double-digit accuracy gains over ANN compression baselines ported to SNNs. We further report a synaptic-operation-based energy proxy and a calibration-size ablation, demonstrating robust performance under sub-one-sample-per-class calibration.
\end{abstract}


\section{Introduction}

Spiking Neural Networks (SNNs) have garnered significant attention as a promising power-efficient alternative to traditional Artificial Neural Networks (ANNs). They are composed of spiking neurons that communicate via discrete spikes, analogous to their biological counterparts. This sparse, spike-based neural dynamics offers benefits such as low computational power and compatibility with temporal data~\cite{snnsurvey2022}, as well as high power efficiency when implemented on neuromorphic chips like True North~\cite{truenorth2015}, Loihi 1 and 2~\cite{orchard2021efficientneuromorphicsignalprocessing}, or SpiNNaker~\cite{spinnaker2014}. However, as these chips have limited computing power and memory, finding ways to improve SNNs' efficiency on neuromorphic hardware is an important research question~\cite{neurmorphiceffiency}. In this work, we focus on model compression for SNNs, specifically post-training pruning and quantization.

\noindent\textbf{SNN pruning techniques.}\quad Several methods for pruning SNNs have emerged in the past five years. UPF~\cite{shi2024towards} finds that aggressive unstructured pruning ($\geq 90\%$ sparsity) can yield 10x efficiency increase. \cite{guo_unsupervised_2020} proposed an unsupervised online pruning technique for SNNs. \cite{kim2022exploringlotterytickethypothesis} explored the lottery ticket hypothesis (LTH) for SNN and used it as the basis for their unstructured weight pruning method. However, most state-of-the-art (SOTA) SNN pruning methods either require pruning during training or adopt an iterative pruning approach, in which multiple pruning and training iterations are performed. Such approaches are computationally expensive, especially for pre-trained, deep spiking neural networks like SEW-ResNet 152 \cite{fang2022deepresiduallearningspiking}, and Spiking Transformers \cite{yao2023spikedriventransformer, zhou2022spikformerspikingneuralnetwork}. With the adoption of transformer architectures for SNNs, the scale of future state-of-the-art SNN models could make retraining costs prohibitively large. Therefore, a more computationally efficient pruning algorithm is urgently needed for the deployment of future next-generation SNNs.

\noindent\textbf{SNN quantization techniques}\quad Efforts to apply quantization to SNNs on neuromorphic hardware have recently gained traction. There are two main categories in neural network quantization: Quantization-Aware Training (QAT) and Post Training Quantization (PTQ). QAT allows models to adapt to the quantization through retraining, and there have been several articles written on QAT for SNNs, such as \cite{lui2021hessianawarequantizationspiking, QAT2, wei2025qp}. However, PTQ for SNNs is still relatively underdeveloped, especially compared to its ANN counterparts, where the desire to reduce the size of Large Language Models (LLMs) to run on smaller or even edge computing hardware has inspired many algorithms, such as AdaQuant \cite{jhunjhunwala2021adaptivequantizationmodelupdates}, Optimal Brain Quantizer (OBQ) \cite{frantar2023optimalbraincompressionframework}, and GPTQ \cite{frantar2023gptqaccurateposttrainingquantization}. 

\noindent\textbf{Other efficiency improvement techniques.}\quad Some other notable methods of improving SNN efficiency include knowledge distillation~\cite{knowledgedistillation1, knowledgedistillation2}, where a larger model is used to train a smaller model, spike rate reduction~\cite{spikerate1, spikerate2, spikerate3} and hardware-specific optimizations~\cite{hardwareopt1, hardwareopt2}. Due to space limitations, we do not cover these methods in depth here; they are orthogonal and can be used to complement pruning/quantization, as seen in recent works~\cite{shymyrbay2023lowprecisionquantizationawaretraining, chowdhury2021spatiotemporalpruningquantizationlowlatency}. 

We discovered that the naive application of ANN one-shot post-training compression methods, such as Optimal Brain Compression~\cite{frantar2023optimalbraincompressionframework}, to SNNs yields suboptimal results due to a disconnect between ANN methods' objective functions and SNNs' layer-wise output spike trains. To address this research gap, we derived an innovative objective function based on the Van Rossum Distance (VRD)~\cite{vanrossum} of output spike trains, referred to as the surrogate membrane potential (SMP). By statistically estimating SMP's Hessian from a calibration dataset, SBC maintains OBC's computational cost while providing tighter compression guarantees.

Based on SMP, we introduce Spiking Brain Compression, a layer-wise compression algorithm that leverages SMP to compress pre-trained SNNs in one shot. We validate SBC on pruning and quantization tasks with neuromorphic benchmarks and large-scale static datasets, demonstrating efficient, accurate compression of SNNs.

The main contributions of this work can be summarized as follows: 
\begin{itemize}
    \item We formalize a VRD–based loss for layer-wise SNN compression and derive an efficient Hessian that can be sampled from a small calibration dataset.
    \item We develop \emph{Spiking Brain Compression (SBC)}: a one-shot second-order SNN compression algorithm that supports modern SNN architectures, including deep spiking CNNs and spiking Transformers.
    \item SBC demonstrates state-of-the-art one-shot post-training SNN compression across neuromorphic and large-scale static benchmarks (up to ImageNet), showing scalability by compressing SEW-ResNet152~\cite{fang2022deepresiduallearningspiking} and Spike-Driven Transformers~\cite{yao2023spikedriventransformer}. 
    \item A calibration-size ablation showed stable performance under data-scarce, sub-one-sample-per-class calibration.
\end{itemize}

\section{Preliminary}

\subsection{Leaky-Integrate-and-Fire (LIF) neuron}

We use the discrete LIF neuron model described in \cite{fang2021incorporatinglearnablemembranetime}, with $V_{reset}=0$. As summarized in Eqs.~\ref{eq:ut}–\ref{eq:vt}, $U[t]$ and $V[t]$ represent the neuron membrane potential after neuron dynamics, and after the trigger of a spike at time t, respectively. $I[t]$ denotes the external input current. $S[t]$ denotes the output spike train. $\tau_m$ is the membrane time constant, which governs the rate of decay for the membrane potential.
\begin{align}
U[t] &= \left(1 - \frac{1}{\tau_m} \right)V[t-1] + \frac{1}{\tau_m} I[t] \label{eq:ut} \\
S[t] &= \Theta(U[t] - V_{th}) \label{eq:st} \\
V[t] &= U[t](1 - S[t]) \label{eq:vt}
\end{align}

\subsection{Optimal Brain Surgeon} 
Neural network compression methods that use second-order approximation are extensions of the OBS framework \cite{OBS}, which considers the problem of accurately compressing a well-trained dense neural network. We briefly review the OBS rules for selecting the next weight to compress $w_p$, and the optimal compensation $\delta_p$, given the expected Hessian $\boldsymbol{H}$ of loss $\mathcal{L}$:
\begin{equation}
    w_p = argmin_{w_p} \frac{c_{w_p}^2}{[\boldsymbol{H}^{-1}]_{pp}}, \delta_p = -\frac{c_{w_p}}{[\boldsymbol{H}^{-1}]_{pp}} \cdot [\boldsymbol{H}^{-1}]_{:p} \label{eq:obs}
\end{equation}
Here, $c_{w_p}$ denotes the changes to weight p (for pruning $c_{w_p} = w_p$, for quantization $c_{w_p} = w_p -\hat{w_p}$ where $\hat{w_p}$ is the nearest quantized expression of $w_p$). $[\boldsymbol{H}^{-1}]_{pp}$ denotes the $p^{th}$ diagonal of inverse Hessian, $[\boldsymbol{H}^{-1}]_{:p}$ is its $p^{th}$ column.

\subsubsection{Optimal Brain Compression Hessian Matrix Update} 
As a weight is pruned, its Hessian matrix $\boldsymbol{H'}=\boldsymbol{H_{-p}}$, where $\boldsymbol{H}_{-p}$ represents matrix $\boldsymbol{H}$ with row p and column p removed. However, $\boldsymbol{H_{-p}^{-1}} \neq \boldsymbol{(H^{-1})_{-p}}$, and calculating the inverse Hessian of the remaining matrix at each pruning iteration can incur a high computational cost. Building upon OBS, Optimal Brain Compression~\cite{frantar2023optimalbraincompressionframework} introduced a straightforward method to obtain the inverse of a Hessian matrix without row p and column p, using the Sherman-Morrison formula:
\begin{equation}
    \boldsymbol{H}_{-p}^{-1} = (\boldsymbol{H^{-1}-\frac{1}{[H^{-1}]_{pp}}H_{:,p}^{-1}H_{p,:}^{-1}})_{-p}
    \label{eq:sherman-morrison}
\end{equation}

\begin{figure*}[t]
  \centering
  \small
  \begin{tikzpicture}
    \matrix (m) [%
      matrix of nodes,
      nodes={layer, rotate=90},
      column sep=1.0cm,
      row sep=0cm
    ] {
      Conv2d & BatchNorm2d & LIFNeuron & Conv2d & LIFNeuron & Flatten & Linear & LIFNeuron \\
    };

    \foreach \i in {1,...,7} {
      \draw[arrow] (m-1-\i) -- (m-1-\the\numexpr\i+1\relax);
    }

    \draw[decorate, decoration={brace, amplitude=5pt, raise=8pt}]
      (m-1-1.north east) -- node[brace label, yshift=18pt] {Module 1} (m-1-3.south east);

    \draw[decorate, decoration={brace, amplitude=5pt, raise=8pt}]
      (m-1-4.north east) -- node[brace label, yshift=18pt] {Module 2} (m-1-5.south east);

    \draw[decorate, decoration={brace, amplitude=5pt, raise=8pt}]
      (m-1-7.north east) -- node[brace label, yshift=18pt] {Module 3} (m-1-8.south east);

  \end{tikzpicture}
  \caption{Example SNN modules: each ends with a LIF layer and includes preceding linear/conv/BN layers.}
  \label{fig:module-demo}
\end{figure*}

\section{Methodology}

\noindent\textbf{Module-wise compression.}\quad\label{sec:module_wise_compression} Since small current errors accumulate through LIF dynamics, we measure distortion at the output spike train. To achieve this, we compress SNNs \emph{module-wise}: Every module comprises Linear/Conv(+BatchNorm) $\rightarrow$ LIF layers; BatchNorm and Conv can be folded into a single linear map, so each module becomes Linear(W) $\rightarrow$ LIF. (Figure \ref{fig:module-demo} visualizes the module concept in an SNN). For an input spike tensor \(X\in\{0,1\}^{T\times d_{\text{in}}}\) and weights \(W\in\mathbb{R}^{d_{\text{in}}\times d_{\text{out}}}\), the module produces a spike train \(S=f(X,W)\in\{0,1\}^{T\times d_{\text{out}}}\). We treat each module as an independent compression problem. 

\noindent\textbf{Problem setup.}\quad The goal of module-wise compression is to find a "compressed" $W$, which we denote $\hat{W}$, that performs as similarly to the original weights as possible. More formally, the compressed weights $\hat{W}$ should minimize some loss function $\mathcal{L}$ while satisfying some constraints on $\hat{W}$, expressed as $\mathcal{C}(\hat{W}) > C$:
\begin{equation}
    argmin_{\hat{W}} E_X[\mathcal{L}(f(X, W), f(X, \hat{W}))]\text{, s.t. }\mathcal{C}(\hat{W}) > C \label{eq:objective_function}
\end{equation}
The expectation is approximated with $N$ calibration samples. 

\subsection{Spiking Brain Compression (SBC)}

This paper identifies an appropriate loss function $\mathcal{L}$ and computes its expected Hessian $\boldsymbol{H}$ for general SNNs. We refer to this framework as Spiking Brain Compression.

\subsubsection{Loss function $\mathcal{L}$}

Let $\hat{S}=f(X, \hat{W})$ and $S=f(X, W)$. Due to the discrete nature of $S$ and $\hat{S}$, using the squared L2 Norm on $S - \hat{S}$ ignores the time distances between modified and original spikes. To address this, we used the squared VRD between the pre- and post-compression output spike trains as the loss function. Here we represent the VRD decaying exponential kernel $k(t)$ as a function of time constant $\tau$:
\begin{equation}
    k[t] = \begin{cases}
         (1-\frac{1}{\tau_m})^{t}\frac{1}{\tau_m} & \text{if $t \geq 0$} \\
         0 & \text{if $t < 0$}
        \end{cases}
\end{equation}
The loss can then be expressed as:
\begin{equation}
\begin{split}
\label{eq:vrd}
    \mathcal{L}(W) &= VRD(S, \hat{S}) \\
    &= ||S(t)*k(t) - \hat{S}(t)*k(t)||_2^2 \\
    &= ||MS-M\hat{S}||_2^2
\end{split}
\end{equation}
Where the square matrix $M$ of size $d_{time} \times d_{time}$ denotes the convolution matrix of kernel $k(t)$ over $d_{time}$ timesteps. Since the output spike train of an LIF neuron depends solely on its own external input current, and $k(t)$ operates as a convolution along the time dimension, the VRD of each LIF neuron only depends on the weights feeding into the neuron. Consequently, the loss in Eq.~\ref{eq:vrd} can be expressed as the sum of the square VRD of each neuron in the layer.
\begin{equation}
    ||MS - M\hat{S}||_2^2 = \sum_{j=1}^{d_{out}}||MS_{:, j} - M\hat{S}_{:,j}||_2^2 \label{eq:perchannel}
\end{equation} 
Therefore, the Hessian for the synaptic connections of each LIF neuron (i.e., each column of the weight matrix \(W\)) can be computed independently, as there are no inter-neuron dependencies. In the following section, we analyze the loss function of an individual neuron, with lower case $s$, $\hat{s}$, and $w$ representing the spiketrains and synaptic connections of an individual LIF neuron: $\mathcal{L} = ||Ms-M\hat{s}||_2^2=||Mf(X, w)-Mf(X, \hat{w})||_2^2$.

\subsubsection{Surrogate Membrane Potential (SMP)}

Here we provide SMP, an approximation of the exact Hessian that is computationally efficient and empirically validated across both small and large SNNs. We first provide the Hessian $\mathcal{H}$ of the loss function $\mathcal{L}$ under the OBS framework. Detailed derivation can be found in Appendix \ref{appendix:exact_hessian}:
\begin{equation} \label{eq:hessian2}
\begin{split}
    \mathcal{H}_{ij} &= 2(M\frac{\partial s}{\partial u}x_j)^TM\frac{\partial s}{\partial u}x_i + 2(Ms-M\hat{s})^TMx_i\frac{\partial^2 s}{\partial u^2}x_j \\
\end{split}
\end{equation}
Where $s$ and $u$ stand for the layer spike train and layer exact membrane potential, and $\frac{\partial s}{\partial u}$ is the $d_{times} \times d_{times}$ Jacobian, which we name $h'$, and $\frac{\partial^2 s}{\partial u^2}$ we rename $h''$. 

We observe that the spike at time t $s[t] = \Theta(u[t]-V_{th})$ (Eq. \ref{eq:st}) only depends on the membrane potential at time t. This means $h'$ and $h''$ are diagonal matrices. Inspired by surrogate gradient \cite{neftci2019surrogategradientlearningspiking}, we replace the non-differentiable Heaviside function $\Theta$ with a differentiable gradient function $g$, thus $h'$ is a diagonal matrix where $h'_{ii} = g(u_{i})$. 

There are many different ways to design a surrogate gradient function $g$, and we intend to explore them in future work. Currently, we found that a constant function $g(u)=c$, works well for layer-wise compression. $g$ can be seen as a \emph{shallow rectangle surrogate gradient} where every $u$ falls in the active window. Since the OBS framework~\cite{OBS} works with the relative magnitude of the Hessian matrix, c cancels out. Without loss of generality, we set $c=1$. Let $\boldsymbol{H} = E_X[H]$ be the expectation of the Hessian over a distribution of input X. With a constant function $g$, we also obtain $h'' = g' = 0$. We now obtain the Hessian for SMP:
\begin{equation}
\label{eq:hessian_surrogate2}
\begin{split}
    \mathcal{H_{SMT}}_{ij} &= 2(Mh'x_j)^TMh'x_i = 2(Mx_j)^TMx_i\\
    \boldsymbol{H}_{SMT} &= E_X[\mathcal{H}_{SMT}] = E_X[2(MX)^TMX]
\end{split}
\end{equation} 
In fact, $\boldsymbol{H}_{SMT}$ is the exact Hessian of the least square form $||MXw - MX\hat{w}||_2^2$, which is the square L2 Loss on the membrane potential of a spiking neuron with the input current $Xw$ in the absence of spikes. 

\subsubsection{The SBC Pruning Algorithm}

Here, we introduce a module-wise unstructured pruning framework for SNNs that utilizes SMP. We start by obtaining a per-module pruning target with Layer-Adaptive sparsity for the Magnitude-based Pruning Score (LAMPS)~\cite{lamps} to determine per-module pruning percentage from a global pruning target. Appendix~\ref{appendix:LAMPS} provides a detailed explanation of LAMPS. LAMPS takes $O(d \cdot log(d))$, where $d$ represents the trainable parameter count in the SNN.

\noindent\textbf{Step 1: Weight ordering.}\quad To efficiently determine the pruning order of weights in each module, we record each weight's loss by performing OBS per neuron with a small batch size $B_{in}$. Each iteration, we take weights with $B_{in}$ smallest loss according to Eq.~\ref{eq:obs} and prune them together, then update the inverse Hessian with Woodbury Matrix Identity~\cite{OBS}. The per-neuron full algorithm is given in Algorithm~\ref{alg:single_pruning}. For small $B_{in}$, this takes $O(\frac{d_{\mathrm{in}}^{3}}{B_{in}})$ time and $O(d_{in})$ space.
\begin{algorithm}[H] 
    \caption{Losses $\mathbf{L}$ for weights $\mathbf{w}$ of neuron with $\mathbf{H}^{-1}=(2\mathbf{(MX)}^{\top}\mathbf{MX})^{-1}$, in $O({d_{\mathrm{in}}^{3}}/{B_{in}})$ time.}
    \label{alg:single_pruning}
    \begin{algorithmic}
    \STATE $\mathbb{M} \gets \{1,\ldots,d_{\mathrm{in}}\}$
    \STATE $\mathbf{L} \gets \{\}$
    \FOR{$i = 1,1+B_{in},\ldots,d_{in}$}
        \STATE $s_p \gets \dfrac{w_p^{2}}{[\mathbf{H}^{-1}]_{pp}} \ \ \forall p\in \mathbb{M}$
        \STATE $\mathbb{P} \gets \text{indices of the $B_{\text{in}}$ smallest } \{s_p\}_{p\in \mathbb{M}}$
        \STATE $\mathbf{L}[p] \gets s_p \ \ \forall p\in\mathbb{P}$
        \STATE $\mathbf{w}_{\mathbb{P}} \gets \mathbf{w}_p \ \ \forall p\in\mathbb{P}$
        \STATE $\mathbf{w} \gets \mathbf{w} - \mathbf{H}^{-1}_{:,\mathbb{P}} ((\mathbf{H}^{-1})_{\mathbb{P}})^{-1} \cdot \mathbf{w}_\mathbb{P}$
        \STATE $\mathbf{H}^{-1} \gets \mathbf{H}^{-1} - \mathbf{H}^{-1}_{:, \mathbb{P}} ((\mathbf{H}^{-1})_{\mathbb{P}})^{-1} \mathbf{H}^{-1}_{\mathbb{P}, :}$
        \STATE $\mathbb{M} \gets \mathbb{M} \setminus \{\mathbb{P}\}$
    \ENDFOR
    \end{algorithmic}
\end{algorithm}

In actuality, we can batch $B_{out}$ neurons in parallel, given the available GPU resources. This produces a loss for each weight in the module, which we then sort and create a mask $\mathbf{M}$ of pruned weights according to the module pruning target. This takes $O(\frac{d_{out}}{B_{out}} \cdot \frac{d_{in}^3}{B_{in}})$ time and $O(B_{out} \cdot d_{in}^2)$ space. 

\noindent\textbf{Step 2: Weight pruning.}\quad With a mask $\mathbf{M}$, we can directly update weights of a neuron according to its local set of weights to remove $\mathbb{P} = \mathbf{M}_{:,i}$ via the group OBS formula $\delta_{\mathbb{P}} = -\mathbf{H}^{-1}_{:, \mathbb{P}} ((\mathbf{H^{-1}})_\mathbb{P})^{-1}\mathbf{W}_{:,i}$. This takes $O(\frac{d_{out}}{B_{out}}\cdot d_{in}^3)$ time and $O(B_{out} \cdot d_{in}^2)$ space, but it is generally faster than the weight ordering step. 
\begin{algorithm}[H] 
\caption{SBC Pruning, with SNN $\mathcal{M}$, calibration data $\mathcal{X}$, model sparsity target $s \in [0, 1]$ }
\label{alg:sbc_pruning}
\begin{algorithmic}
\STATE $\mathbb{M} \gets \text{modules}(\mathcal{M})$ \COMMENT{get set of modules}
\FORALL{$\mathbf{m} \in \mathbb{M}$} 
    \STATE $s_m = \text{LAMPS}(\mathcal{M}, s, \mathbf{m})$ \COMMENT{module target}
    \STATE $\mathbb{X} \gets \text{module } \mathbf{m} \text{ input data from } \mathcal{X}, \mathcal{M}$
    \STATE $\mathbf{H} = \frac{2}{|\mathbb{X}|}\sum_{X\in\mathbb{X}}(MX)^TMX$
    \STATE $\mathbf{W} \gets \text{weights of } \mathbf{m}$
    \STATE $\mathbf{L} \gets \textbf{Algorithm 1}(\mathbf{W}, \mathbf{H}^{-1})$
    \STATE $\mathbf{M} \gets \text{indices of } |\mathbf{W}|\cdot s_\mathbf{m} \text{ smallest }\mathbf{L}$
    \FOR{$i = 1,\ldots d_{out}$}
        \STATE $\mathbb{P} \gets \mathbf{M_{:,i}} $
        \STATE $\delta_{\mathbb{P}} \gets -\mathbf{H}^{-1}_{:, \mathbb{P}} ((\mathbf{H^{-1}})_\mathbb{P})^{-1}\mathbf{W}_{:,i}$
        \STATE $\mathbf{W}_{:, i} \gets \mathbf{W}_{:, i} + \delta_{\mathbb{P}}$
    \ENDFOR
\ENDFOR
\end{algorithmic}
\end{algorithm}
\noindent\textbf{Complexity.}\quad To summarise, per-module SBC has space complexity $O(B_{out}\cdot d_{\text{in}}^{2})$ and time complexity $O(\frac{d_{out}}{B_{out}} \cdot \frac{d_{in}^3}{B_{in}})$, with $d_{in}, d_{out}$ refer to the input and output dimension of the linearized parameterized layer in the module. Batch sizes $B_{in} $ and $ B_{out}$ are adjusted to balance time and space constraints based on hardware availability.

\noindent\textbf{Modern SNN architectures.}\quad SBC can also handle compression of modern SNNs like Spiking ResNets \cite{fang2022deepresiduallearningspiking} and Spiking Transformer \cite{zhou2022spikformerspikingneuralnetwork, yao2023spikedriventransformer} architectures. Appendix \ref{appendix:deepsnn} provides implementation details. The intuition is to treat each module separately, and when two different parameterized layers feed into the same spiking neuron layer, to concatenate the parameterized layers' weights and inputs.

\begin{table*}[t]
  \caption{Datasets and SNN architectures used for compression. $\tau_m$ is the membrane time constant of the LIF neuron layers, and T is the number of timesteps. Note that the SEW-ResNet family uses Integrate-and-Fire (IF) neurons, which are equivalent to LIF neurons with $\tau_m=\infty$.}
  \label{datasnntable}
  \centering
  \begin{tabular}{llllllll}
    \toprule
    Dataset     & Architecture  & Input Shape & $\tau_m$  & T & Accuracy(\%) \\
    \midrule
    N-MNIST         & 2FC       & 34x34x2 & 2.0 & 100 & 98.31 \\
    CIFAR10-DVS     & 4Conv+2FC & 64x64x2 & 2.0 & 20  & 71.50 \\
    DVS128-Gesture  & 5Conv+2FC & 128x128x2 & 2.0 & 20 & 95.14 \\
    ASL-DVS         & 4Conv+1FC & 180x240x2 & 2.0  & 30 & 96.53 \\
    CIFAR-100        & VGG16SNN  & 32x32x3 & 1.33 & 5 & 71.05 \\
    ImageNet        & SEW-ResNets & 256x256x3 & $+\infty$ & 4 & 63.12-69.18 \\
    ImageNet        & Spike-Driven Transformer & 256x256x3 & 2.0 & 4 & 72.28-77.07 \\
    \bottomrule
  \end{tabular}
\end{table*} 

\subsubsection{The SBC Quantization Algorithm} 
Quantization is essential to the practical deployment of SNNs on neuromorphic hardware. Mainstream neuromorphic hardware stores parameters in low precision, as part of the hardware design choice and/or to save space. For example, SpiNNaker 2 (optimized for 8- and 16-bit), Tianjic (8-bit), Loihi 1 (9-bit), Loihi 2 (up to 32, but default to 8-bit), and BrainScaleS-2 (6-bit)~\cite{spinnaker2014, tianjic8998338, orchard2021efficientneuromorphicsignalprocessing, pehle2022brainscales2acceleratedneuromorphichybrid}.

Furthermore, prior work~\cite{liu2023quantization} demonstrated that naive quantization of SNNs performs poorly on neuromorphic hardware. This underscores the need for a robust post-training quantization algorithm, such as ours, to enable large-scale SNN deployment.

The SBC quantization algorithm largely follows GPTQ~\cite{frantar2023gptqaccurateposttrainingquantization}, replacing the layer-wise Hessian with $\mathbf{H}_{SMP}$. It has per-module space complexity $O(d_{in}^2)$ and time complexity $O(max\{d_{out} \cdot d_{in}^2, d_{in}^3\})$.

\subsection{Comparing SBC with OBC} \label{justification}

A naive implementation of OBC for SNNs would use a loss function directly on the linearized layer's output. It has the Hessian:
\begin{equation} \label{eq:hessian_obc}
    \boldsymbol{H}_{OBC} = E_X[\mathcal{H}_{OBC}] = E_X[2X^TX]
\end{equation}
Recall Eq.\ref{eq:hessian_surrogate2}, $M\frac{\partial s}{\partial u} = Mh'$ always preserves the lower triangle structure (for spike trains with spikes) because $M$ is lower triangular, and $h'$ is diagonal. This means the simplification $Mh'=I$, i.e. $\boldsymbol{H}_{OBC}$, is typically weaker than keeping $M$, i.e. $\boldsymbol{H}_{SMT}$.

\section{Experiments}

To explore how SBC works in practice, we performed pruning and quantization on models trained with neuromorphic and static datasets. 
SNNs can leverage unstructured pruning to reduce compute~\cite {shi2024towards} on neuromorphic hardware, unlike ANNs on GPUs. 
For this reason, the experiments are focused on pruning, while we conducted quantization experiments only on N-MNIST, CIFAR10-DVS, and DVS128-Gesture. 

The specifications of each model are listed in Table~\ref{datasnntable}. SNN implementation is based on the SpikingJelly library~\cite{spikingjelly}. Experiments were conducted on a single Nvidia A4500 (20GB) GPU.

Four models were trained on neuromorphic datasets N-MNIST, DVS128 Gesture, DVS-CIFAR10, and ASL-DVS. N-MNIST was trained on the Adam optimizer with $lr=0.001$ and a maximum of 200 epochs. DVS-CIFAR10 and DVS128-Gestures were trained on the Adam optimizer with Cosine Annealing $lr=0.001$, and max epoch 512. ASL-DVS was trained with the SGD optimizer and Cosine Annealing, with $lr=0.001$ for 90 epochs. N-MNIST, DVS128-Gesture, and ASL-DVS maintained their original size, whereas DVS-CIFAR10 was downsized due to the long training time. 

The Spiking VGG16 model was chosen for the CIFAR-100 dataset. It was trained on SGD optimizer for 300 epochs, with steps at 150 and 225, using an initial learning rate of $lr_{initial} = 0.1$. SEW-ResNet family and Spike-Driven Transformer (SDT) model checkpoints were retrieved from their respective GitHub repositories.

\subsection{Pruning Result and Analysis}

\subsubsection{SBC vs One-shot Post-Training SNN Pruning Algorithms}

\begin{figure*}
  \centering
    \includegraphics[width=\linewidth,trim=0 6 0 6,clip]{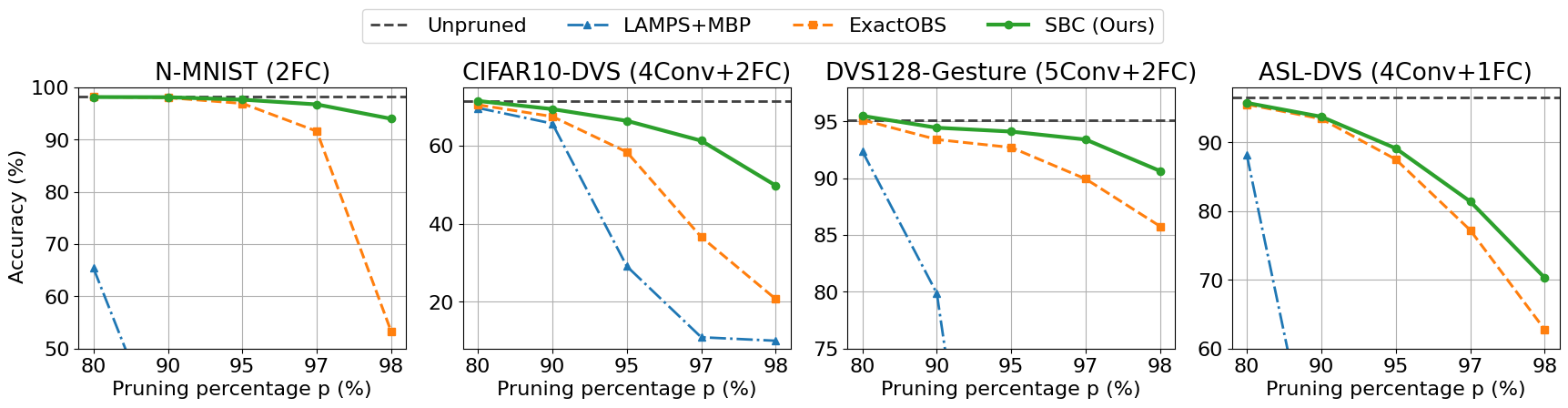}
  \vspace{2pt} 
    \includegraphics[width=\linewidth,trim=0 6 0 6,clip]{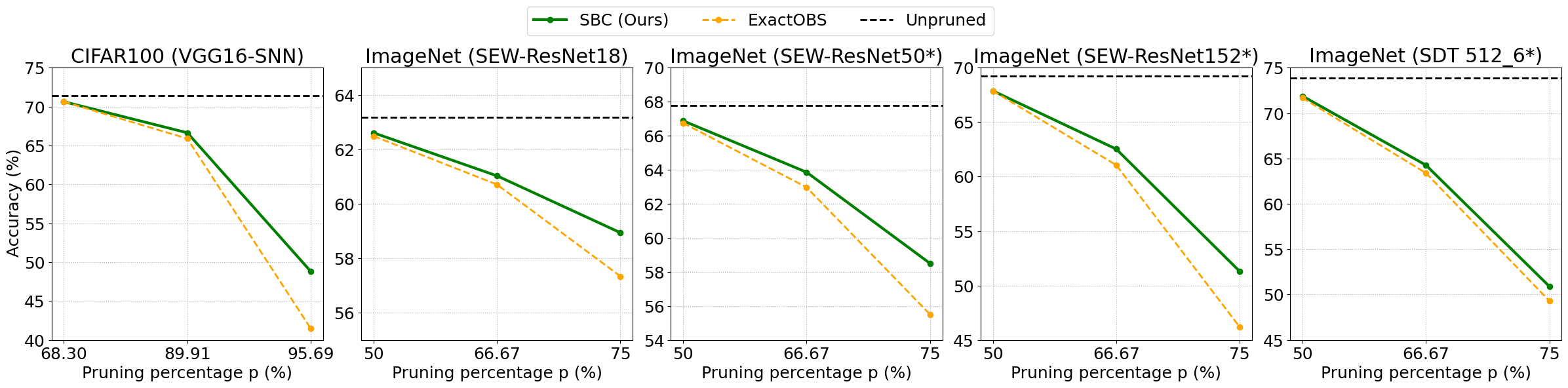}
  \caption{Neuromorphic (top row) and static (bottom row) dataset pruning results, post-training one-shot SNN pruning algorithms. (*) models are the largest SNNs pruned to date. Tables: Appx.~\ref{appendix:result_tables}}
  \label{fig:prunetable}
\end{figure*}

We compared the performance of the naive magnitude-based pruning (MBP) shown in LAMPS, the naive ExactOBS (OBC's pruning algorithm) implementation on SNN, and SBC at different sparsity levels across all datasets. The accuracies are recorded in Fig.~\ref{fig:prunetable}. Note that all three post-training one-shot methods pruned the same proportion of weights from each module according to LAMPS; the difference is in how each module prunes its weights. 

Across neuromorphic benchmarks, SBC substantially outperforms both ExactOBS and LAMPS+MBP as sparsity increases. In particular, while all methods start from the same unpruned accuracy, SBC maintains high accuracy under extreme pruning (e.g., N-MNIST: -1.59\% at 97\% sparsity compared to ExactOBS’s -45.07\%; DVS128: -1.74\% at 97\% sparsity compared to ExactOBS's -9.38\%), and shows similar robustness on CIFAR10-DVS and ASL-DVS where competing methods degrade more rapidly. In short, SBC preserves performance at low sparsity and extends the usable sparsity range relative to other one-shot post-training methods.

The static-dataset results mirror this trend: SBC has the largest gains compared to ExactOBS at very high sparsities (e.g., CIFAR-100 VGG16-SNN: +7.47\% at 75\% sparsity). On ImageNet across SEW-ResNet variants and Spike-Driven Transformer, SBC consistently matches or improves ExactOBS, demonstrating that SBC is the SOTA one-shot post-training pruning algorithm for SNNs.

\subsubsection{SBC vs PAT Algorithms}

\begin{figure*}
    \centering
    \includegraphics[width=\linewidth]{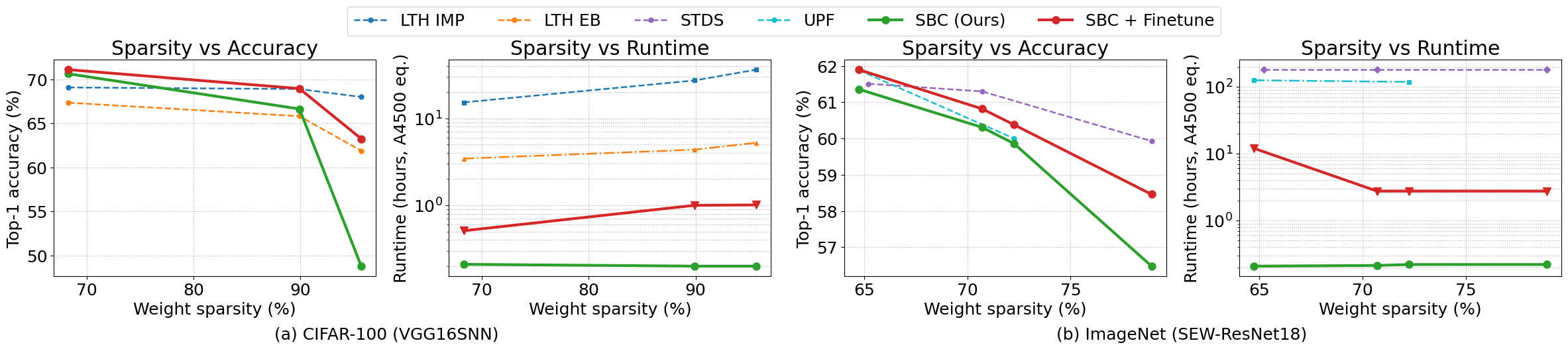}
    \caption{Large static dataset pruning results, SBC (Ours) vs PAT methods. Tables: Appx.~\ref{appendix:result_tables}}
    \label{fig:sbcvpat}
\end{figure*}

We compare SBC (with and without post-pruning fine-tuning) against representative PAT baselines, including LTH~\cite{kim2022exploringlotterytickethypothesis}, UPF~\cite{shi2024towards}, and STDS~\cite{stds}; accuracy and wall-clock pruning times are reported in Fig.~\ref{fig:sbcvpat}. Across CIFAR-100 and ImageNet, SBC reaches competitive accuracy, and adding a short fine-tuning phase further closes the gap to PAT methods while incurring minimal additional computational cost.

Concretely, on CIFAR-100 at 68.30\% and 89.91\% sparsity, SBC improves top-1 accuracy by 2.02\% and 0.06\% versus the IMP found in LTH, while reducing pruning time by roughly two orders of magnitude. SBC also outperforms LTH Early Bird methods across the evaluated sparsity targets while running 5–7× faster. On ImageNet, SBC matches or approaches UPF/STDS accuracy but requires substantially less time (three orders of magnitude faster for the pruning step); with fine-tuning, SBC surpasses UPF’s accuracy but remains slightly below STDS. These results indicate that SBC is a cost-effective alternative to full PAT pipelines: it delivers strong, deployable sparsity with far lower compute and provides flexibility to trade a small additional cost for further accuracy gains.

\subsubsection{Large SNNs Pruning}

Crucially, the lower computation cost enables SBC to prune large SNNs such as SEW ResNet152 and Spike-Driven Transformer 512\_6, when PAT methods become too costly. To the best of our knowledge~\cite{kim2022exploringlotterytickethypothesis, stds, shi2024towards, wei2025qp}, this experiment is the first time SNNs with more than 34 layers have been pruned, the first spiking Transformer trained on ImageNet-scale dataset being pruned, and the first SNNs trained on an ImageNet-scale dataset with more than 19 layers have been pruned (Kim's experiment on Spiking ResNet-34~\cite{kim2022exploringlotterytickethypothesis} was trained on Tiny-ImageNet~\cite{Le2015TinyIV}). To the best of our knowledge, SEW-ResNet152 is also the largest and deepest SNN model that has been subjected to unstructured pruning by any algorithm to date. This reaffirms SBC as the SOTA one-shot post-training compression method for SNNs. 


\begin{table*}[hbt!]
  \centering
  \begin{threeparttable}
  \caption{Neuromorphic datasets quantization result}
  \label{quanttable_maintext}
  \begin{tabular}{%
      l   
      l   
      c   
      c   
      c   
      c   
      c   
      c   
    }
    \toprule
    \thead{Dataset} &
    \thead{Architecture} &
    \thead{Time step} &
    \thead{Uncompressed \\ Acc. (\%)} &
    \thead{Top-1 Acc.\\ \textbf{SBC(Ours)} (\%)} &
    \thead{Top-1 Acc.\\ExactOBS (\%)} &
    \thead{Top-1 Acc.\\RTN (\%)} &
    \thead{Bit width} \\
    \midrule
\multirow{3}{*}{N-MNIST}
  & 2FC       & 100 & 98.31 & 98.14 & \textbf{98.16} & 97.58 & 4-bit \\
  &           &     &       & \textbf{97.81} & 97.33 & 62.26 & 3-bit \\
  &           &     &       & \textbf{92.40} & 64.29 & 20.34 & 2-bit \\
\midrule
\multirow{3}{*}{CIFAR10-DVS}
  & 4Conv+2FC & 20  & 71.50 & 69.96 & \textbf{70.30} & 66.40 & 4-bit \\
  &           &     &       & \textbf{69.64} & 67.56 & 52.70 & 3-bit \\
  &           &     &       & \textbf{60.70} & 59.40 & 20.50 & 2-bit \\
\midrule
\multirow{3}{*}{DVS128-Gesture}
  & 5Conv+2FC & 20  & 95.14 & \textbf{87.43} & 86.67 & 87.15 & 4-bit \\
  &           &     &       & \textbf{81.88} & 81.25 & 78.47 & 3-bit \\
  &           &     &       & \textbf{64.79} & 62.78 & 53.82 & 2-bit \\
\bottomrule
  \end{tabular}
  \end{threeparttable}
\end{table*}

In conclusion, the SBC pruning method consistently outperforms current one-shot post-training pruning algorithms for SNNs, achieving one-shot post-training pruning SOTA for SNNs. It is competitive and sometimes even exceeds the accuracy of the SOTA SNN iterative pruning method~\cite{kim2022exploringlotterytickethypothesis}, with 1 to 3 orders of magnitude of time savings. 

\subsection{Quantization Result and Analysis}

We compared SBC's quantization accuracy with a baseline method and vanilla implementation of GPTQ~\cite{frantar2023gptqaccurateposttrainingquantization}, the state-of-the-art quantization derivative of OBC, on SNN. The quantization experiment used the same trained models and sample data as the pruning experiment. The models were quantized to 4, 3, and 2 bit widths. 

This work used the GPTQ framework for quantization experiments. GPTQ quantizes each linear/convolution ANN layer according to standard uniform per-channel symmetric quantization on the min-max grid, similar to~\cite{dettmers2022llmint88bitmatrixmultiplication}. We chose a symmetric quantization grid for the low computing cost when deploying to neuromorphic hardware. The weights are then quantized in the ascending diagonal inverse Hessian order, resulting in a quantization order from largest to smallest loss on an equal-spacing quantization grid, according to Eq.~\ref{eq:obs}. The quantization grids and weight order are the same as GPTQ. It is notable that SBC quantization can be applied to arbitrary quantization grids. 

The baseline method, which \textbf{R}ounds each weight \textbf{T}o its \textbf{N}earest quantized target value, is called RTN. Note that all three methods (RTN, vanilla GPTQ, and SBC) share the same quantization grid, which means all three methods have precisely the same set of quantization target values to round to, and the difference in accuracy only comes from the choice of target. The result of the quantization experiment can be found in Table~\ref{quanttable_maintext}. Each quantization result is computed as the average of 5 runs using randomly selected sample datasets.

SBC's performance is generally better than that of RTN and GPTQ. It can quantize the CIFAR10-DVS model to 3-bit precision with a 1.86\% drop in accuracy. However, SBC does not consistently outperform RTN and GPTQ in higher-precision quantization for CIFAR10-DVS and DVS128-Gesture. This is caused by the weight compensation, which pushes later-quantized weights to more extreme values. These extreme values are outside of the range of the quantization grid. Nevertheless, SBC achieves the best quantization result among the three methods at the lowest bit representation tested (2bit), outperforming the next-best method by 28.11\%, 1.30\%, and 2.01\%, respectively.

\subsection{Efficiency Estimation with SOPs}
We use synaptic operations (SOPs)~\cite{shi2024towards} as a proxy metric to estimate the energy consumption of an SNN on neuromorphic hardware. This is supported by~\cite{basu2022spiking}, a detailed review of existing neuromorphic hardware. In Table~\ref{sops_table}, we provide average SOPs for 4Conv-2FC trained on CIFAR10-DVS, VGGSNN-16 trained on CIFAR-100, and SEW-ResNet18 trained on ImageNet at different sparsities. These results demonstrated SBC's effectiveness in reducing SOPs while maintaining model accuracy. 
\begin{table}[h]
\caption{Synaptic operations per forward pass at different sparsity levels.}
\label{sops_table}
\centering
\begin{tabular}{l c c c}
    \toprule 
    \thead{Dataset / Model} & 
    \thead{Sparsity (\%)} & 
    \thead{SOPs (M)} &
    \thead{Accuracy (\%)} \\
    \midrule
    \multirow{2}{*}{\makecell[l]{CIFAR10-DVS\\4CONV+2FC}}
               & 0    & 119.2 & 71.50 \\
              & 90  & 25.51 & 69.40 \\
    \midrule
    \multirow{2}{*}{\makecell[l]{CIFAR-100\\VGGSNN-16}}
               & 0    & 107.95 & 71.05\\
               & 89.91  & 35.85 & 68.96 \\
    \midrule
    \multirow{2}{*}{\makecell[l]{ImageNet\\SEW-ResNet18}}
               & 0    & 1134.40 & 63.18\\
               & 72.26  & 697.24 & 60.38\\
    \bottomrule
\end{tabular}
\end{table}

\subsection{Ablation on Calibration Data Size}
We evaluated SBC under varying calibration-set sizes on three models: 4Conv2FC on CIFAR-10-DVS, VGGSNN-16 on CIFAR-100, and SEW-ResNet18 on ImageNet. For each setting, we drew three independent stratified calibration datasets of the indicated size, applied SBC using only these samples, and reported the mean and standard deviation of the resulting test accuracy. Calibration size in Table~\ref{calibrationtable} denotes the absolute size of the calibration dataset.
\begin{table}[h]
\caption{Classification accuracy at different calibration dataset sizes for each dataset/model pair.}
\label{calibrationtable}
\centering
\begin{tabular}{l c c}
    \toprule 
    \thead{Dataset / Model} & 
    \thead{Calibr. size} & 
    \thead{Accuracy (\%)} \\
    \midrule
    \multirow{4}{*}{\makecell[l]{CIFAR10-DVS\\4CONV+2FC}}
               & 10    & 49.64 ($\pm$1.27)  \\
              & 90    & 61.90 ($\pm$0.21)  \\
              & 900   & 63.84 ($\pm$0.49)  \\
              & 9000  & 64.04 ($\pm$0.35)  \\
    \midrule
    \multirow{3}{*}{\makecell[l]{CIFAR-100\\VGGSNN-16}}
               & 100    & 62.19 ($\pm$0.43) \\
               & 1000   & 65.80 ($\pm$0.07) \\
               & 10000  & 66.42 ($\pm$0.17) \\
    \midrule
    \multirow{4}{*}{\makecell[l]{ImageNet\\SEW-ResNet18}}
               & 100    & 49.35 ($\pm$0.19) \\
               & 1000   & 55.06 ($\pm$0.12) \\
               & 10000  & 55.89 ($\pm$0.11) \\
               & 50000  & 55.89 ($\pm$0.04) \\
    \bottomrule
\end{tabular}
\end{table}

These results show that SBC already achieves strong accuracy on the neuromorphic dataset with only 90 calibration samples per class, and on the static datasets with as few as 1–10 samples per class. Interestingly, for SEW-ResNet18 on ImageNet, even 100 calibration images (0.1 images/class) yield competitive performance. In all but one case, the standard deviations are $\leq 0.5\%$, demonstrating SBC's robustness against the size of the calibration set.

\section{Discussion} \label{sec:limiations}

We have identified two limitations of SBC that will be a significant part of the future work. First, in quantization experiments, we observed the out-of-bounds problem with the min-max quantization grid. A study on the relationship between the choice of the quantization grid and quantization performance is needed. Secondly, we aim to explore more sophisticated surrogate gradients $g'$ and $M$, or even a general $d_{time} \times d_{times}$ matrix in place of $M$ in the current $\boldsymbol{H}_{SMP}$. These matrices can be adjusted depending on each neuron's properties that are not expected to change during compression, such as spike rates. However, we would like to note that a naive implementation of custom Hessian matrices for each output would have high space complexity $\Theta(d_{out} \cdot d_{in}^2)$. In the conv2 layer of the last blocks in the SEW-ResNet152, which has linearized $d_{in}=4608$ and $d_{out}=512$, the Hessian would take up $512 \times 4608^2 \times 4bit = 43.5$GB of space at fp32. As SNN architectures scale, this cost grows quickly, making a lower-complexity strategy essential for practical compression use cases.

\section{Conclusion}

In conclusion, this paper proposes a one-shot, efficient, post-training compression framework for SNNs, utilizing a second-order approximation of the per-layer spike train loss to dynamically compress and compensate for the compression. Through empirical analysis, this paper shows SBC's effectiveness in compressing SNNs trained on neuromorphic and static datasets, outperforming current SOTA on SNN post-training one-shot methods, and being competitive with the accuracy of iterative retraining compression methods while providing a 1-2 orders of magnitude decrease in compression time. We expect this work to pave the way for the efficient compression of very large SNNs, such as deep SNNs and spiking Transformers, for low-power, edge computing, and robotic use cases.

\section*{Impact Statement}

This paper presents work whose goal is to advance the field of Machine
Learning. There are many potential societal consequences of our work, none of
which we feel must be specifically highlighted here.


\bibliography{references}
\bibliographystyle{icml2026}

\newpage
\appendix
\onecolumn

\section{Deriving exact Hessian for individual LIF neuron}
\label{appendix:exact_hessian}
We derive the first-order derivative of loss function $\mathcal{L}$ against weight $w_i$:

\begin{equation} \label{eq:L_first_order}
\begin{split}
    \frac{\partial\mathcal{L}}{\partial w_i} &= \frac{\partial\mathcal{L}}{\partial s} \frac{\partial s}{\partial w_i} 
    = \frac{||Ms-M\hat{s}||_2^2}{\partial s} \frac{\partial s}{\partial w_i} \\
    &= 2(Ms-M\hat{s})^TM \frac{\partial s}{\partial w_i}
\end{split}
\end{equation}

As is customary in OBS framework, with a well-trained neural network, we assume $\nabla_{w_j}\mathcal{L} = 0$ for all weights $w_i$. Now we can get the second-order derivative: 

\begin{equation} \label{eq:hessian1}
\begin{split}
    \mathcal{H}_{ij} &=\frac{\partial^2 \mathcal{L}}{\partial w_i\partial w_j}\\ 
    &= \frac{\partial}{\partial w_j}(2(Ms-M\hat{s})^TM \frac{\partial s}{\partial w_i}) \quad \text{(by \eqref{eq:L_first_order}})\\
    &= 2[\frac{\partial}{\partial w_j}(a^Tb)] \quad \text{where } a=(Ms-M\hat{s}), b=M\frac{\partial s}{\partial w_i}\\
    &= 2[(\frac{\partial a}{\partial w_j})^Tb + a^T(\frac{\partial b}{\partial w_j})] \quad \text{(product rule)}\\
    &= 2[(\frac{\partial (Ms-M\hat{s})}{\partial w_j})^TM\frac{\partial s}{\partial w_j} + (Ms-M\hat{s})^T\frac{\partial^2 Ms}{\partial w_i \partial w_j}] \quad \text{expand a and b}\\
    &= 2[(\frac{\partial (Ms)}{\partial w_j})^TM\frac{\partial s}{\partial w_j} + (Ms-M\hat{s})^TM\frac{\partial^2 s}{\partial w_i \partial w_j}] \\
    &= 2(M\frac{\partial s}{\partial w_j})^TM \frac{\partial s}{\partial w_i} + 2(Ms-M\hat{s})^TM \frac{\partial^2 s}{\partial w_i \partial w_j} \\
\end{split}
\end{equation}

We can further derive $\frac{\partial s}{\partial w_i}$ and $\frac{\partial^2 s}{\partial w_i \partial w_j}$:

\begin{equation}
\begin{split}
    \frac{\partial s}{\partial w_i} &= \frac{\partial s}{\partial u}\frac{\partial u}{\partial w_i} = \frac{\partial s}{\partial u}x_i \\
    \frac{\partial^2 s}{\partial w_i \partial w_j} &= \frac{\partial}{\partial w_j}(\frac{\partial s}{\partial u}x_i) = x_i \frac{\partial}{\partial w_j}(\frac{\partial s}{\partial u}) = x_i [\frac{\partial}{\partial u}(\frac{\partial s}{\partial u})] \frac{\partial u}{\partial w_j}=x_i\frac{\partial^2 s}{\partial u^2}x_j \\
\end{split}
\end{equation}

Where $x_i, x_j$ denotes time-varying pre-synaptic spike trains to synaptic connections $w_i$ and $w_j$, i.e. $x_i = X_{:, i}$. We can now further simplify the exact Hessian Eq. \ref{eq:hessian1}: 

\begin{equation} \label{eq:hessian2}
\begin{split}
    \mathcal{H}_{ij} &= 2(M\frac{\partial s}{\partial u}x_j)^TM\frac{\partial s}{\partial u}x_i + 2(Ms-M\hat{s})^TMx_i\frac{\partial^2 s}{\partial u^2}x_j \\
\end{split}
\end{equation}

\section{Derivation on how to apply SBC to various Spiking ResNets and Spiking Transformer architectures}
\label{appendix:deepsnn}

\subsection{Spiking ResNets} 

\paragraph{Shortcut handling in Spiking ResNets}
We derive the SBC loss with surrogate membrane potential (SMP) for two popular Spiking ResNet variants: Spiking-Element-Wise (SEW) ResNet~\cite{fang2022deepresiduallearningspiking} and SpikingResNet~\cite{spikingresnet}.
In SEW ResNet, the shortcut follows the spiking layer; from SBC’s perspective, the two branches are already independent modules and can be pruned separately.

In SpikingResNet, the shortcut precedes the spiking layer. Thus the external current to the next spiking layer is the sum of the convolution output and the shortcut output. Consider one neuron with linearized convolution output \(X_1 w_1\), shortcut output \(X_2 w_2\), and surrogate membrane-potential operator \(M\).
The shortcut loss is
\begin{align}
L_{\text{shortcut}}
&= \mathbb{E}_{X}\!\left[ \left\| M\!\left(X_1 w_1 + X_2 w_2\right) - M\!\left(X_1 \hat{w}_1 + X_2 \hat{w}_2\right) \right\|_2^2 \right] \\
&= \mathbb{E}_{X}\!\left[ \left\| M\!\left(\,[\,X_1 \;\; X_2\,]\!
\begin{bmatrix} w_1 \\ w_2 \end{bmatrix}\right)
- M\!\left(\,[\,X_1 \;\; X_2\,]\!
\begin{bmatrix} \hat{w}_1 \\ \hat{w}_2 \end{bmatrix}\right) \right\|_2^2 \right].
\end{align}
Hence we may concatenate the inputs and weights of the last convolution and the shortcut in each SpikingResNet block, forming a new module with input
\(X = [\,X_1 \;\; X_2\,]\) and weights \(W = \begin{bmatrix} w_1 \\ w_2 \end{bmatrix}\).
Compression is then performed on this module. If the shortcut has no trainable parameters, we simply mask out the weights associated with \(X_2\) before compression.
\emph{Notation:} \([\,X_1 \;\; X_2\,]\) denotes horizontal concatenation; \(\begin{bmatrix} w_1 \\ w_2 \end{bmatrix}\) denotes vertical concatenation.

\subsection{Spiking Transformers}

This section examines the applicability of \textbf{SBC} to popular \emph{spiking–transformer} architectures, and we find that it extends cleanly to \mbox{Spikformer}~\cite{zhou2022spikformerspikingneuralnetwork}, Spikformer~V2~\cite{zhou2024spikformerv2joinhigh}, and the Spike-Driven Transformer (SDT)~\cite{yao2023spikedriventransformer}.

\paragraph{Spikformer family.}
In Spikformer, the weights of the Spiking Self-Attention (SSA) layers
$(W_Q, W_K, W_V)$ belong to a  
\(\text{Linear}\!\rightarrow\!\text{BN}\!\rightarrow\!\text{Spiking}\) block and can therefore be treated as \emph{three} SBC modules.  
Each MLP block follows  
\(\text{Linear}\!\rightarrow\!\text{BN}\!\rightarrow\!\text{Spiking}\!\rightarrow\!\text{Linear}\!\rightarrow\!\text{BN}\!\rightarrow\!\text{Spiking}\),  
giving \emph{two} additional modules.  
Spikformer~V1 uses Spiking Patch Splitting (SPS), whereas V2 employs a Spiking Convolution Stem (SCS); both are sequential convolutions that fit the SBC module-wise compression definition in Section~\ref{sec:module_wise_compression}.

\paragraph{Spike-Driven Transformer (SDT).}
In SDT, the shortcut precedes the spiking layer, so the external current is the sum of the previous BN output and the shortcut. Let the main-path output be \(XW\), the shortcut output \(U\), and \(M\) the surrogate membrane-potential operator.  
For a single spiking layer, the shortcut loss is
\begin{align}
L_{\text{shortcut}}
 &= \mathbb{E}_{X}\!\bigl[\|M(XW+U)-M(X\hat{W}+U)\|_2^2\bigr] \notag\\
 &= \mathbb{E}_{X}\!\bigl[\|MXW+MU-MX\hat{W}-MU\|_2^2\bigr] \notag\\
 &= \mathbb{E}_{X}\!\bigl[\|MXW-MX\hat{W}\|_2^2\bigr].
\end{align}
Hence, the shortcut can be ignored during compression, allowing all SDT modules to be treated exactly as in Spikformer.

\section{Experiment Implementation details}
\label{appendix:experimentdetails}

Note that we provide anonymised code for all experiments.

\subsection{Static Models}
We use the Spiking VGG16 model implemented by ~\cite{kim2022exploringlotterytickethypothesis} for CIFAR100 experiments. We used the pre-trained SEW-ResNet and Spike-Driven Transformer families of models for our ImageNet pruning experiments. 

\subsection{Neuromorphic Models}

The detailed architectures of the models applied to each neuromorphic dataset are given in the figures \ref{fig:nmnist-model}, \ref{fig:cifar10dvs-model}, \ref{fig:dvs128-gesture-model}, \ref{fig:asldvs-model}. All LIF neuron layers use the surrogate function $Atan()$, time constant $ \tau=2.0$. Note the downsizing of CIFAR10-DVS dataset to 64x64, due to the computation limitation of our hardware. 

\begin{figure}[htbp]
  \centering
  \begin{tikzpicture}
    \matrix (m) [%
      matrix of nodes,
      nodes={layer, rotate=90, text width=3.0cm},
      column sep=1.0cm,
      row sep=0cm
    ] {
      Input 34x34x2xT & Flatten layer & {Linear layer \\ $2312\to256$} & {LIF Neuron layer} & {Linear layer \\ $256\to10$} & LIF Neuron layer\\
    };

    \foreach \i in {1,...,5} {
      \draw[arrow] (m-1-\i) -- (m-1-\the\numexpr\i+1\relax);
    };
    
  \end{tikzpicture}
  \caption{N-MNIST 2FC architecture}
  \label{fig:nmnist-model}
\end{figure}

\begin{figure}[htbp]
  \centering
  \begin{tikzpicture}
    \matrix (m) [%
      matrix of nodes,
      nodes={layer, rotate=90, text width=3.0cm},
      column sep=0.6cm,
      row sep=0cm
    ] {
      Input 128x128x2xT & {Conv2d layer \\ kernel 3x3 \\ 128 channels} & BatchNorm2d & LIF neuron layer & {Max Pooling layer \\ kernel 2x2} & Flatten & {Linear layer \\ $2048\to500$} & LIF neuron layer & {Linear layer \\ $500\to110$} & LIF neuron layer & Voting Layer\\
    };

    \foreach \i in {1,...,10} {
      \draw[arrow] 
      (m-1-\i) -- (m-1-\the\numexpr\i+1\relax);
    }

    \draw[decorate, decoration={brace, amplitude=5pt, raise=8pt}]
      (m-1-2.north east) -- node[brace label, yshift=18pt] {$\times 5$} (m-1-5.south east);

  \end{tikzpicture}
  \caption{DVS128-Gesture 5Conv-2FC architecture}
  \label{fig:dvs128-gesture-model}
\end{figure}

\begin{figure}[htbp]
  \centering
  \begin{tikzpicture}
    \matrix (m) [%
      matrix of nodes,
      nodes={layer, rotate=90, text width=3.0cm},
      column sep=0.6cm,
      row sep=0cm
    ] {
      Input 64x64x2xT & {Conv2d layer \\ kernel 3x3 \\ 128 channels} & BatchNorm2d & LIF neuron layer & {Max Pooling layer \\ kernel 2x2} & Flatten & {Linear layer \\ $2048\to500$} & LIF neuron layer & {Linear layer \\ $500\to110$} & LIF neuron layer & Voting Layer\\
    };

    \foreach \i in {1,...,10} {
      \draw[arrow] (m-1-\i) -- (m-1-\the\numexpr\i+1\relax);
    }

    \draw[decorate, decoration={brace, amplitude=5pt, raise=8pt}]
      (m-1-2.north east) -- node[brace label, yshift=18pt] {$\times 4$} (m-1-5.south east);

  \end{tikzpicture}
  \caption{CIFAR10-DVS 4Conv-2FC architecture}
  \label{fig:cifar10dvs-model}
\end{figure}

\begin{figure}[htbp]
  \centering
  \begin{tikzpicture}
    \matrix (m) [%
      matrix of nodes,
      nodes={layer, rotate=90, text width=3.5cm},
      column sep=0.6cm,
      row sep=0cm
    ] {
      Input 180x240x2xT & {Conv2d layer \\ kernel 5x5 \\ $12\to32\to48\to48$ \\ channels} & BatchNorm2d & LIF neuron layer & {Max Pooling layer \\ kernel 2x2} & Flatten & {Linear layer \\ $6720\to24$} & LIF neuron layer\\
    };

    \foreach \i in {1,...,7} {
      \draw[arrow] (m-1-\i) -- (m-1-\the\numexpr\i+1\relax);
    }

    \draw[decorate, decoration={brace, amplitude=5pt, raise=8pt}]
      (m-1-2.north east) -- node[brace label, yshift=18pt] {$\times 4$} (m-1-5.south east);

  \end{tikzpicture}
  \caption{ASL-DVS 4Conv-1FC architecture}
  \label{fig:asldvs-model}
\end{figure}

\subsection{Pruning Experiments Details}
\label{appendix:LAMPS}
The pruning percentage for each prunable layers in the module is pre-determined with LAMPS \cite{lamps}. Here is a summary:

\paragraph{LAMPS} Let the weights of layer \(l\) be flattened into
\(\mathbf{W}^{(l)}\in\mathbb{R}^{N_l}\), and let
\(\pi_l\) be the permutation that orders them in non‐decreasing magnitude:
\[
\bigl|W^{(l)}_{\pi_l(1)}\bigr|
\;\le\;
\bigl|W^{(l)}_{\pi_l(2)}\bigr|
\;\le\;
\cdots
\;\le\;
\bigl|W^{(l)}_{\pi_l(N_l)}\bigr|.
\]
Then for \(u=1,\dots,N_l\), the \emph{LAMP score} of the \(u\)-th smallest weight in layer \(l\) is
\begin{equation}\label{eq:lamp-score}
\mathrm{LAMP}\bigl(l,\pi_l(u)\bigr)
\;=\;
\frac{\bigl(W^{(l)}_{\pi_l(u)}\bigr)^{2}}
     {\sum_{v=u}^{N_l}\bigl(W^{(l)}_{\pi_l(v)}\bigr)^{2}}.
\end{equation}
To prune to a global sparsity level \(S\), collect all scores
\(\{\mathrm{LAMP}(l,\pi_l(u))\}\) across layers, and remove
\(\lfloor S\sum_l N_l\rfloor\) weights from each layer. 

The weights in each module are then pruned in the order suggested by LAMPS scores; however, the exact weights pruned and compensations are determined by the SBC pruning algorithm. The exact implementation can be found in the supplementary file osbc\_prune.py

\subsection{Quantization}

The quantization grid is chosen as a symmetric per-channel fixed-grid of width $\Delta_c$. This means the maximum absolute value of a channel of weights equals the maximum value of the quantization grid. The symmetric quantization grid enables easy scalar conversion between compressed and original quantized values. 

\begin{equation}
    \hat{w_c} = w_{c_{quantized}} \times \Delta_c, w_{c_{quantized}} \in \{2^{n-1}, ...,-1, 0, 1, ... 2^{n-1}-1\}
\end{equation}

For an n-bit quantization grid for channel c. The implementation of this maximum value grid is directly lifted from the GPTQ \cite{frantar2023gptqaccurateposttrainingquantization} implementation on GitHub. It can be found in the supplementary file quantizer.py. 

The quantization experiment was run five times at each quantization level, with five different seeds for the PyTorch and NumPy random number generators used to select a sample dataset from the training dataset. The exact seeds are available in modelutils.py. The implementation for OBSC quant can be found in osbc\_quant.py in the supplementary material.

\section{Experiment Result Tables}

\label{appendix:result_tables}
\begin{table}[hbt!]
  \caption{Neuromorphic dataset pruning result}
  \label{prunetable}
  \centering
  \begin{tabular}{%
      l       
      l       
      *{6}{c} 
    }
    \toprule
    & & \multicolumn{5}{c}{\thead{Accuracy(\%)}} \\
    \cmidrule(lr){3-8}
    \thead{Setting} 
      & \thead{Method} 
      & \thead{$p=0\%$} 
      & \thead{$80.00\%$} 
      & \thead{$90.00\%$} 
      & \thead{$95.00\%$} 
      & \thead{$97.00\%$} 
      & \thead{$98.00\%$} \\
    \midrule
    \multirow{3}{*}{\makecell[l]{N-MNIST\\2FC}}
      & LAMPS+MBP           & 98.31  & 65.39  & 29.40 & 35.13 & 26.94 & 25.60 \\
      & ExactOBS            & 98.31  & \textbf{98.16}  & 98.01 & 96.91 & 91.59 & 53.24 \\
      & \textbf{SBC(Ours)}  & 98.31 & 98.13  & \textbf{98.10} & \textbf{97.63} & \textbf{96.72} & \textbf{93.97} \\
    \midrule
    \multirow{3}{*}{\makecell[l]{CIFAR10-DVS\\4Conv+2FC}}
      & LAMPS+MBP   & 71.50  & 69.70  & 65.70 & 29.10 & 10.90 & 10.00 \\
      & ExactOBS    & 71.50  & 70.50  & 67.50 & 58.40 & 36.60 & 20.70 \\
      & \textbf{SBC(Ours)}  & 71.50  & \textbf{71.50}  & \textbf{69.40} & \textbf{66.40} & \textbf{61.30} & \textbf{49.80} \\
    \midrule
    \multirow{3}{*}{\makecell[l]{DVS128-Gesture\\5Conv+2FC}}
      & LAMPS+MBP   & 95.14  & 92.36  & 79.86 & 35.13 & 9.03 & 8.33 \\
      & ExactOBS    & 95.14  & 95.13  & 93.40 & 92.7 & 89.93 & 85.76 \\
      & \textbf{SBC(Ours)}  & 95.14  & \textbf{95.48}  & \textbf{94.44} & \textbf{94.10} & \textbf{93.40} & \textbf{90.63} \\
    \midrule
    \multirow{3}{*}{\makecell[l]{ASL-DVS\\4Conv+1FC}}
      & LAMPS+MBP   & 96.53  & 88.11  & 36.48 & 11.76 & 7.19 & 6.08 \\
      & ExactOBS    & 96.53  & 95.51  & 93.41 & 87.52 & 77.23 & 62.78 \\
      & \textbf{SBC(Ours)}  & 96.53  & \textbf{95.73}  & \textbf{93.72} & \textbf{89.12} & \textbf{81.40} & \textbf{70.30} \\
    \midrule
  \end{tabular}
\end{table}

\begin{table}[hbt!]
  \centering
  \begin{threeparttable}

  \caption{Static dataset pruning results: SBC vs.\ ExactOBS}
  \label{tab:static_dataset_obcsbc}
  \begin{tabular}{%
      l   
      l   
      c   
      c   
      c   
      c   
      c   
      c   
    }
    \toprule
    \thead{Dataset} &
    \thead{Arch.} &
    \thead{Time\\step} &
    \thead{Top-1 Acc.\\ \textbf{SBC(Ours)} \\(\%)} &
    \thead{Top-1 Acc.\\ExactOBS \\(\%)} &
    \thead{Weight\\Sparsity \\(\%)} &
    \thead{Time (h)\\A4500 Eq.} &
    \thead{Calib. size\\(\% train)} \\
    \midrule
\multirow{3}{*}{CIFAR100}
  & VGG16-SNN & 5 & 71.40 & 71.40 & 0 & - & - \\
  &           &   & \textbf{70.63} & \textbf{70.63} & 68.30 & 0.21 & 20\% \\
  &           &   & \textbf{66.44} & 65.89 & 89.91 & 0.20 & 20\% \\
  &           &   & \textbf{48.93} & 41.46 & 95.69 & 0.20 & 20\% \\
\midrule
\multirow{5}{*}{ImageNet}
  & SEW-ResNet18  & 4 & 63.18 & 63.18 & 0 & - & - \\
  &               &   & \textbf{62.61} & 62.49 & 50.00 & 0.208 & 2\% \\
  &               &   & \textbf{61.03} & 60.71 & 66.67 & 0.214 & 2\% \\
  &               &   & \textbf{58.94} & 57.34 & 75.00 & 0.220 & 2\% \\
\cmidrule(lr){2-8}
  & SEW-ResNet50†  & 4 & 67.78 & 67.78 & 0 & - & - \\
  &               &   & \textbf{66.88} & 66.75 & 50.00 & 0.346 & 2\% \\
  &               &   & \textbf{63.86} & 62.97 & 66.67 & 0.344 & 2\% \\
  &               &   & \textbf{58.50} & 55.50 & 75.00 & 0.341 & 2\% \\
\cmidrule(lr){2-8}
  & SEW-ResNet152† & 4 & 69.26 & 69.26 & 0 & - & - \\
  &               &   & \textbf{67.88} & \textbf{67.88} & 50.00 & 0.783 & 2\% \\
  &               &   & \textbf{62.53} & 61.06 & 66.67 & 0.808 & 2\% \\
  &               &   & \textbf{51.30} & 46.19 & 75.00 & 0.723 & 2\% \\
    \bottomrule
  \end{tabular}

  \begin{tablenotes}[flushleft]\footnotesize
    \item[*] \textbf{Bolded} entries represent best accuracies in sparsity class
    \item[†] Deepest SNN models pruned to date. No existing PAT methods have pruned them.
  \end{tablenotes}
  \end{threeparttable}
\end{table}

\begin{table}[hbt!]
  \centering
  \begin{threeparttable}

  \caption{Static dataset pruning results, SBC compared with Pruning-Aware-Training methods}
  \label{tab:static_dataset_patptc}
  \begin{tabular}{%
      l  
      l  
      l  
      c  
      c  
      c  
      c  
      c  
    }
    \toprule
    \thead{Dataset \\ / Model} &
    \thead{Time\\step} &
    \thead{Pruning\\Method} &
    \thead{Top-1\\Acc.\,(\%)} &
    \thead{Weight\\Sparsity\\\,(\%)} &
    \thead{Time (h)\\A4500 Eq.} &
    \thead{Calib. \\size\\(\% train)} &
    \thead{Finetune \\ epochs} \\
    \midrule
\multirow{9}{*}{\makecell[l]{CIFAR100\\VGG16-\\SNN}}
  & \multirow{9}{*}{5}
  & \multirow{3}{*}{LTH IMP}
      & 69.08 & 68.30 & 15.23 & -- & --  \\
  & & & 68.90 & 89.91 & 27.22 & -- & --  \\
  & & & \textbf{68.00} & 95.69 & 36.22 & -- & --  \\
  \cmidrule(lr){3-8}
  & & \multirow{3}{*}{LTH EB}
      & 67.35 & 68.30 &  3.44 & -- & --  \\
  & & & 65.82 & 89.91 &  4.36 & -- & --  \\
  & & & 61.90 & 95.69 &  5.24 & -- & --  \\
  \cmidrule(lr){3-8}
  & & \multirow{1}{*}{\textbf{SBC (Ours)}}
      & 70.63 & 68.30 & \textbf{0.21} & 20\% & --\\
  & &  & 66.44 & 89.91 & \textbf{0.20} & 20\% & --\\
  & &  & 48.93 & 95.69 & \textbf{0.20} & 20\% & --\\
  \cmidrule(lr){3-8}
  & & \multirow{1}{*}{\textbf{SBC + Finetune}}
      & \textbf{71.10(+0.47†)} & 68.30 & 0.51 & 20\% & 25.0\\
  & &  & \textbf{68.96(+2.52†)} & 89.91 & 0.81 & 20\% & 50.0\\
  & &  & 63.34(+14.41†) & 95.69 & 0.81 & 20\% & 50.0\\
\midrule
\multirow{10}{*}{\makecell[l]{ImageNet\\SEW-\\ResNet18}}
  & \multirow{10}{*}{4}
  & \multirow{2}{*}{STDS}
      & 61.51 & 65.21 & 180   & -- & --  \\
  & & & \textbf{61.30} & 70.71 & 180   & -- & --  \\
  & & & \textbf{59.93} & 78.92 & 180   & -- & --  \\
  \cmidrule(lr){3-8}
  & & \multirow{2}{*}{UPF}
      & 61.89 & 64.74 & 125   & -- & --  \\
  & & & 60.00 & 72.26 & 118   & -- & --  \\
  \cmidrule(lr){3-8}
  & & \multirow{1}{*}{\textbf{SBC (Ours)}}
      & 61.36 & 64.74 & \textbf{0.208} & 2\% & -- \\
  & &  & 60.31 & 70.71 & \textbf{0.213} & 2\% & -- \\
  & &  & 59.86 & 72.26 & \textbf{0.220} & 2\% & -- \\
  & &  & 56.48 & 78.92 & \textbf{0.220} & 2\% & -- \\
  \cmidrule(lr){3-8}
  & & \multirow{1}{*}{\makecell[l]{\textbf{SBC + Finetune}}}
      & \textbf{61.90(+0.54†)} & 64.74 & 12.00 & 2\% & 5.0 \\
  & &  & 60.82(+0.51†) & 70.71 & 2.75 & 2\% & 1.0 \\
  & &  & \textbf{60.38(+0.52†)} & 72.26 & 2.75 & 2\% & 1.0 \\
  & &  & 58.46(+1.98†) & 78.92 & 2.75 & 2\% & 1.0 \\
    \bottomrule
  \end{tabular}

  \begin{tablenotes}[flushleft]\footnotesize
    \item[*] \textbf{Bolded} entries represent best accuracies/shortest compression time in sparsity class
    \item[†] Accuracy improvements compared to pre-finetuned, SBC pruned models
  \end{tablenotes}
  \end{threeparttable}
\end{table}

\begin{table}[hbt!]
  \centering
  \begin{threeparttable}

  \caption{ImageNet pruning on Spike-Driven Transformer: SBC vs.\ ExactOBS}
  \label{tab:sdt_imagenet_obcsbc}
  \begin{tabular}{%
      l   
      l   
      c   
      c   
      c   
      c   
      c   
      c   
    }
    \toprule
    \thead{Dataset} &
    \thead{Arch.} &
    \thead{Time\\step} &
    \thead{Top-1 Acc.\\ \textbf{SBC (Ours)} \\(\%)} &
    \thead{Top-1 Acc.\\ExactOBS \\(\%)} &
    \thead{Weight\\Sparsity \\(\%)} &
    \thead{Time (h)\\A4500 Eq.} &
    \thead{Calib. size\\(\% train)} \\
    \midrule
\multirow{4}{*}{ImageNet}
  & SDT 6\_512 & 4 & \textbf{73.89} & \textbf{73.89} & 0.00 & --    & --  \\
  &            &   & \textbf{71.88} & 71.68          & 50.00 & 0.212 & 1\% \\
  &            &   & \textbf{64.27} & 63.40          & 66.67 & 0.204 & 1\% \\
  &            &   & \textbf{50.91} & 49.31          & 75.00 & 0.206 & 1\% \\
    \bottomrule
  \end{tabular}

  \begin{tablenotes}[flushleft]\footnotesize
    \item[*] \textbf{Bolded} entries represent best accuracies in each sparsity class.
  \end{tablenotes}
  \end{threeparttable}
\end{table}

\end{document}